\documentclass[10pt,twocolumn,letterpaper]{article}

\usepackage[pagenumbers]{cvpr} % To force page numbers, e.g. for an arXiv version

\definecolor{cvprblue}{rgb}{0.21,0.49,0.74}
\usepackage[accsupp]{axessibility}
\usepackage[pagebackref,breaklinks,colorlinks,allcolors=cvprblue]{hyperref}
\usepackage{adjustbox}
\usepackage{url}            % simple URL typesetting
\usepackage{booktabs}       % professional-quality tables
\usepackage{amsfonts}       % blackboard math symbols
\usepackage{nicefrac}       % compact symbols for 1/2, etc.
\usepackage{microtype}      % microtypography
\usepackage{xcolor}         % colors
\usepackage{enumitem}
\usepackage{bm}
\usepackage{wrapfig}
\usepackage{amsmath}
\usepackage{multicol}
\usepackage{multirow}
\usepackage{subcaption}
\usepackage{xcolor,colortbl}
\usepackage{dsfont}

\definecolor{yellow}{rgb}{1, 1, 0.7}
\definecolor{orange}{rgb}{1, 0.85, 0.7}
\definecolor{red}{rgb}{1, 0.7, 0.7}
\definecolor{normalred}{rgb}{1, 0, 0}

\def\paperID{12070} % *** Enter the Paper ID here
\def\confName{CVPR}
\def\confYear{2025}

\title{MonoInstance: Enhancing Monocular Priors via Multi-view Instance Alignment for Neural Rendering and Reconstruction}

\author{
Wenyuan Zhang$^1$, Yixiao Yang$^1$, Han Huang$^1$, Liang Han$^1$, Kanle Shi$^2$, \\
Yu-Shen Liu$^1$\thanks{The corresponding author is Yu-Shen Liu.} \ , Zhizhong Han$^3$    \\
  School of Software, Tsinghua University, Beijing, China$^1$ \\
  Kuaishou Technology, Beijing, China$^2$ \\
  Department of Computer Science, Wayne State University, Detroit, USA$^3$ \\
  {\tt\small \{zhangwen21,yangyixi21,h-huang20,hanl23\}@mails.tsinghua.edu.cn} \\
  {\tt\small shikanle@kuaishou.com, liuyushen@tsinghua.edu.cn, h312h@wayne.edu} \\
}

\begin{document}
\maketitle

\begin{abstract}
    Monocular depth priors have been widely adopted by neural rendering in multi-view based tasks such as 3D reconstruction and novel view synthesis. However, due to the inconsistent prediction on each view, how to more effectively leverage monocular cues in a multi-view context remains a challenge. Current methods treat the entire estimated depth map indiscriminately, and use it as ground truth supervision, while ignoring the inherent inaccuracy and cross-view inconsistency in monocular priors. To resolve these issues, we propose \textbf{MonoInstance}, a general approach that explores the uncertainty of monocular depths to provide enhanced geometric priors for neural rendering and reconstruction. Our key insight lies in aligning each segmented instance depths from multiple views within a common 3D space, thereby casting the uncertainty estimation of monocular depths into a density measure within noisy point clouds. For high-uncertainty areas where depth priors are unreliable, we further introduce a constraint term that encourages the projected instances to align with corresponding instance masks on nearby views. MonoInstance is a versatile strategy which can be seamlessly integrated into various multi-view neural rendering frameworks. Our experimental results demonstrate that MonoInstance significantly improves the performance in both reconstruction and novel view synthesis under various benchmarks. Project page: \url{https://wen-yuan-zhang.github.io/MonoInstance/}.
    
\end{abstract}

\section{Introduction}
Learning scene representations from multiple posed RGB images is a foundational task in computer vision and graphics~\cite{li2023neuralangelo,yu2022monosdf,barron2021mipnerf,zhu2022nice-slam}, with numerous applications across diverse domains such as virtual reality, robotics and autonomous driving. Bridging the gap between 2D images and 3D representations has become a central challenge in the field. Traditional approaches like Multi-View Stereo (MVS)~\cite{yao2018mvsnet,zhao2023mvpsnet}, address this issue by matching features between adjacent views, followed by dense depth estimation and point cloud fusion. Recent methods tackle this problem more effectively through volume rendering. By learning neural representations, either implicit or explicit ones, like NeRF~\cite{mildenhall2020nerf} and 3D Gaussians~\cite{kerbl20233dgs}, we can conduct volume rendering to rendered these neural representations into images. The rendering results are then supervised by ground truth ones to optimize the neural representations accordingly. Although these methods are capable of generating plausible 3D meshes or novel views~\cite{park2021nerfies,fridovich2023kplanes,wang2021neus}, they struggle to recover fine-grained geometric details. This limitation arises since that the photometric consistency from color images can not ensure perfect geometric clues, which is further complicated by the shape-radiance ambiguity~\cite{zhang2020nerf++}.

To overcome these obstacles, recent solutions typically incorporate monocular priors as additional supervision, such as depths~\cite{yu2022monosdf,song2023darf,zhu2024fsgs} and normals~\cite{wang2022neuris,Dai2024GaussianSurfels,lyu2023learning}. However, the effectiveness of monocular priors becomes a bottleneck hindering the performance of these methods, primarily due to two factors. One is that the predictions from monocular priors are not perfectly accurate due to domain gaps. The other is that monocular priors are inferred independently from each RGB image, leading to geometry inconsistency across different viewpoints. MVS-based methods~\cite{chen2021mvsnerf, wang2023crafting, kaya2022uncertainty} mitigate these issues by deriving the uncertainty through comparing the predicted depths with the projected ones from adjacent views, which is puzzled by view occlusions. While the latest methods~\cite{xiao2024debsdf,chen2024ncsdf} incorporate an additional branch within the rendering framework to predict the uncertainty. However, the uncertainty prediction module in these methods is coupled with the rendering branch, and thus its performance is disturbed by the quality of rendering. 

To resolve these issues, we introduce MonoInstance to enhance monocular priors for neural rendering frameworks by exploring the inconsistency among each instance depths in monocular cues. Our insight builds on the fact that within the same scene, the monocular priors in 3D space will produce depth inconsistency on different views. Hence, when we back-project the depths of the same object from different views into world coordinate system, we can estimate the uncertainty of a 3D point according to the point density in the neighborhood. Specifically, we first segment multi-view images into consistent instances. For each segmented instance, we then back-project and align the multi-view estimated depth values together to create a noisy point cloud. We then evaluate the density of back-projected depth points from each viewpoint within the fused point cloud as the uncertainty measurement, leading to an uncertainty map on each view to highlight the uncertainty area of the instance. For high-uncertainty regions where the priors do not work well, we introduce an additional constraint term, guide the ray sampling, and reduce the weights for inaccurate supervision to infer the geometry and improve rendering details.

We evaluate MonoInstance upon different neural representation learning frameworks in dense-view reconstruction, sparse-view reconstruction and sparse novel view synthesis under the widely used benchmarks. Experimental results show that our method achieves the state-of-the-art performance in various tasks. Our contributions are listed below.

\begin{itemize}
    \item We introduce MonoInstance, which detects uncertainty in 3D according to inconsistent clues from monocular priors on multi-view. Our method is a general strategy to enhance monocular priors for various multi-view neural rendering and reconstruction frameworks.
    \item Based on the uncertainty maps, we introduce novel strategies to reduce the negative impact brought by inconsistent monocular clues and mine more reliable supervision through photometric consistency.
    \item We show our superiority over the state-of-the-art methods using multi-view neural rendering in 3D reconstruction and novel view synthesis on the widely used benchmarks.
\end{itemize}

\section{Related Work}

Neural implicit representations have made a huge
progress in various tasks~\cite{zhou2023uni3d,Ma2025See3D,li2023neuralgf,li2023neaf,li2024learning,zhou2024udiff,zhou2024deepprior,liu2023d-net,wen20223d,xiang2023retro}, which can be learned using different supervision like multi-view~\cite{han2024binocular, huang2025fatesgs, wu2025sparis, zhang2023fast, zhang2024learning} and point
clouds~\cite{ma2021neuralpull, ma2022pcp, ma2022onsurfacepriors, ma2023noise2noise, zhou2023differentiable, chen2025sharpening, chen2024learning, noda2024multipull, noda2025bijective, li2024implicit}. In the following, we focus on reviewing works on learning implicit representations from multi-view.

\noindent\textbf{Neural 3D Reconstruction with Radiance Fields.} Neural Radiance Fields (NeRF) have been a universal technique for multi-view 3D reconstruction. Notable efforts~\cite{wang2021neus, oechsle2021unisurf,yariv2021volsdf} achieve differentiable rendering of neural implicit functions, such as signed distance function~\cite{xiao2024debsdf,zhang2025nerfprior} and occupancy~\cite{oechsle2021unisurf, Hu2023LNI-ADFP}, to infer neural implicit surfaces. Recent approaches introduce various priors as additional supervisions to improve the reconstruction in texture-less areas, such as monocular depth~\cite{yu2022monosdf,xiao2024debsdf,zhang2025nerfprior}, normals~\cite{wang2022neuris, lyu2023learning}, semantic segmentations~\cite{zhou2024manhattanpami,park2023h2o}. More recent methods improve the monocular cues by detecting uncertainties through multi-view projection of depths and normals~\cite{wang2022neuris, wei2021nerfingmvs}, but the projections suffer from view occlusions. Latest methods~\cite{xiao2024debsdf, chen2024ncsdf, tang2024ndsdf} integrate uncertainty estimation within the neural rendering framework, yet the predicted uncertainties are compromised by the rendering quality, especially in complex structures where RGB rendering fails. Moreover, these techniques are specifically designed for indoor scene reconstruction and not applicable across different multi-view neural rendering frameworks. Since there are often only few available views in real-world scenes, some methods are developed for sparse view reconstruction. These methods either are pre-trained on large-scale datasets and finetuned on test scenes~\cite{long2022sparseneus,ren2023volrecon,liang2024retr,na2024uforecon,peng2023gens,peng2024surface}, or leverage monocular priors and cross-view features to overfit a single scene~\cite{huang2024neusurf, wu2023svolsdf}.

\noindent\textbf{Novel View Synthesis with Gaussian Splatting.} Recently, 3D Gaussian Splatting~\cite{kerbl20233dgs} has become a new paradigm in neural rendering due to its fast rendering speed and outstanding performance~\cite{zhou2024diffgs,zhang2024gspull,li2025gaussianudf}. Despite high-quality rendering~\cite{lu2024scaffold,wang2024contextgs}, 3DGS shows poor performance when the number of input views is reduced, due to the overfitted distribution of Gaussians. Recent methods~\cite{zhu2024fsgs,li2024dngaussian,zhang2024corgs,xu2024mvpgs,peng2024structure} tackle this problem by imposing monocular depth priors. However, the priors from pre-trained models often contain significant errors and cannot optimally position the Gaussians. Although monocular depth cues have been widely adopted in multi-view neural rendering and reconstruction frameworks, the uncertainty in depth priors has not been fully explored. To this end, we propose MonoInstance, a universal depth prior enhancement strategy that can seamlessly integrate with various multi-view neural rendering and reconstruction frameworks to improve their performances.

\begin{figure*}[t]
\vspace{-0.0cm}
  \includegraphics[width=\linewidth]{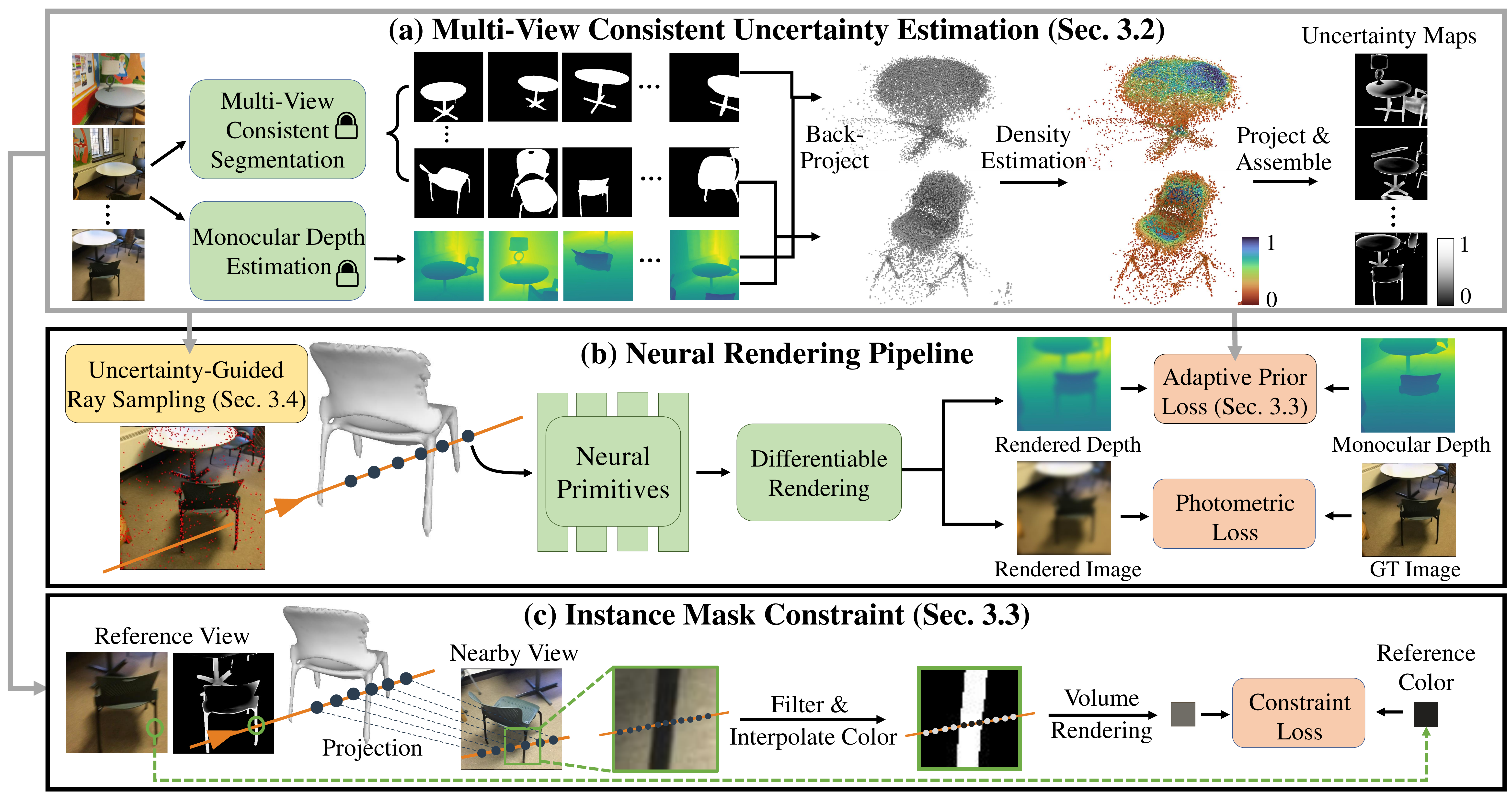}
  \vspace{-0.8cm}
    \caption{Overview of our method. We take multi-view 3D reconstruction through NeRF based rendering as an example. \textbf{(a)} Starting from multi-view consistent instance segmentation and estimated monocular depths, we align the same instance from different viewpoints by back-projecting instance depths into a point cloud. The monocular inconsistent clues across different views become a measurement of density estimation in neighborhood of each point, leading to uncertainty maps (Sec.~\ref{sec:uncertainty}). The estimated uncertainty maps are further utilized in \textbf{(b)} neural rendering pipeline to guide adaptive depth loss, ray sampling (Sec.~\ref{sec:optimization}) and \textbf{(c)} instance mask constraints (Sec.~\ref{sec:silhouette}).}
    \vspace{-0.5cm}
    \label{fig:overview}
\end{figure*}

\section{Method}
Given a set of posed images $\{I_j\}_{j=1}^N$ and the corresponding monocular depth maps $\{D_j\}_{j=1}^N$, we aim to estimate $N$ uncertainty maps $\{U_j\}_{j=1}^N$ according to the inconsistency of monocular depth cues on multi-view images. These uncertainty maps work with our novel strategies to enhance the monocular cues in various neural rendering frameworks to improve the rendering performance and reconstruction accuracy. To achieve this, we introduce a novel scheme to evaluate the uncertainty of 3D points by measuring the point density in a neighborhood. Our novel strategy will use these estimated uncertainty maps to guide the ray sampling, reduce the negative impact brought by the inconsistency, and mine more reliable photometric consistency as a remedy, which thereby enables our method to consistently improve the performance in different neural rendering tasks. An overview of our method is shown in Fig.~\ref{fig:overview}, where we use NeRF-based 3D reconstruction pipeline as an example. See supplementary materials for the differences when applied to 3DGS.
% The implementation differences when applied to 3DGS can be found in Section~\ref{sec:subsec-nvs} and the supplementary materials.

\subsection{Preliminary}

Neural Radiance Fields (NeRF)~\cite{mildenhall2020nerf} and 3D Gaussian Splatting (3DGS)~\cite{kerbl20233dgs} have become paradigms for learning 3D representations from multi-view images. By learning a mapping from 3D positions to densities, NeRF is able to render novel views from given viewpoints using volume rendering,
\vspace{-0.2cm}
\begin{equation}
    \hat{C}(r)=\sum_{i=1}^M \alpha_i T_i c_i, 
    \alpha_i=1-\exp(-\sigma_i\delta_i), T_i=\prod_{k=1}^{i-1}(1-\alpha_k),
\end{equation}
\vspace{-0.03cm}
where $\sigma_i, \delta_i, \alpha_i, c_i$ are the density, sampling interval, opacity and accumulated transmittance at $i$-th sampled point respectively and $\hat{C}(r)$ is the synthesized color of the ray $r$. We can also render depth or normal images in a similar way by accumulating the depth or gradient instead of color,
\vspace{-0.2cm}
\begin{equation}
    \hat{D}(r)=\sum_{i=1}^M \alpha_i T_i t_i, \hat{N}(r)=\sum_{i=1}^M \alpha_i T_i n_i,
\end{equation}
\vspace{-0.03cm}
where $t_i, n_i$ are the sampling distance and gradient of the $i$-th sampled point, respectively. Recent methods extract plausible surfaces from radiance fields by modeling a relationship between SDF and volume density, 
\vspace{-0.2cm}
\begin{equation}
    \sigma(s_i)=
    \begin{cases}
        \frac{1}{2\beta}\ \exp(\frac{-s_i}{\beta}) & \text{if}\  s_i\leq 0 \\
        \frac{1}{\beta}-\frac{1}{2\beta}\ \exp(\frac{s_i}{\beta}) & \text{if}\  s_i>0
    \end{cases},
\end{equation}
\vspace{-0.03cm}
where $\beta$ is a learnable variance parameter and $s_i=\text{SDF}(x_i)$ is the inferred SDF of the sampled point $x_i$. 

Similarly, 3DGS learns 3D Gaussians via differentiable volume rendering for scene modeling,
\vspace{-0.2cm}
\begin{equation}
    \hat{C}(u,v)=\sum_{i=1}^M c_i * o_i * p_i(u,v) \prod_{k=1}^{i-1}(1-o_k * p_k(u,v)), 
\end{equation}
\vspace{-0.03cm}
where $\hat{C}(u,v)$ is the rendered color at the pixel $(u,v)$, $p_i(u,v), c_i, o_i$ denote the Gaussian probability, the color and the opacity of the $i$-th Gaussian projected onto the pixel $(u,v)$, respectively. The neural primitives such as radiance fields and 3D Gaussians can be optimized by minimizing the rendered color and the GT color,
\vspace{-0.05cm}
\begin{equation}
    \mathcal{L}_{color}=\sum_{r\in\mathcal{R}}\Vert\hat{C}(r)-C(r) \Vert_1.
\end{equation}
\vspace{-0.2cm}

\subsection{Uncertainty Estimation from Multi-View Inconsistent Monocular Prior}
\label{sec:uncertainty}
Monocular depth priors have been widely adopted in neural rendering and reconstruction frameworks. However, under the setting of multi-view, the priors struggle to produce consistent results within the same structures from different viewpoints due to the inherent inaccuracy, which makes the optimization even more complex. This issue inspires us to delve into the monocular uncertainty of scene structures from multi-view to provide a more robust prior for neural rendering. To this end, we introduce a novel manner to evaluate uncertainty by point density in a neighborhood after aligning multi-view instances in a unified 3D space. 

\noindent\textbf{Multi-view consistent segmentation.} We first aim to segment every object in the scene to evaluate the uncertainty individually. The reason why we evaluate uncertainty at instance object level is to avoid the impact of object scale on density estimation. Inspired by MaskClustering~\cite{yan2024maskclustering}, we achieve a consistent segmentation across multi-view through a graph-based clustering algorithm. Specifically, we firstly obtain instance segmentation on each image using~\cite{qi2023high}, and then, we connect pairs of instances from different views with an edge to form a graph, if the back-projected depth point clouds of the two instances are close enough in terms of Chamfer Distance. Graph clustering algorithm~\cite{schaeffer2007graph} is then applied to partition the graph nodes into instance clusters. For indoor scenes, based on the assumption that monocular priors in textureless areas are often reliable~\cite{yu2022monosdf,wang2022neuris}, we filter out the background instances and set the uncertainty of the them as zero, using GroundedSAM~\cite{ren2024grounded} as an identification tool. More implementation details can be found in the supplementary materials. 

\begin{figure}[t]
\vspace{-0.0cm}
  \includegraphics[width=\linewidth]{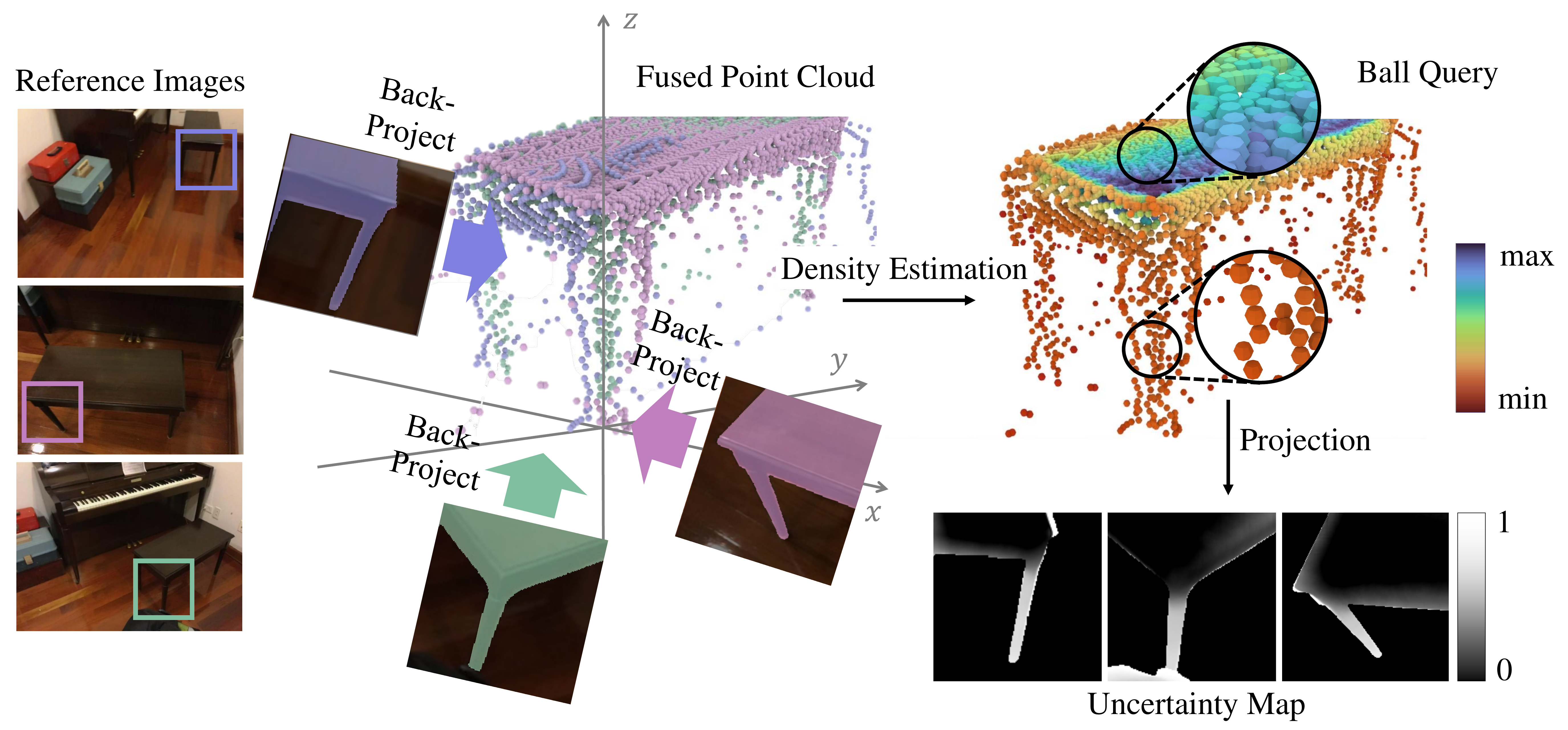}
  \vspace{-0.8cm}
    \caption{Illustration of uncertainty estimation. Areas with inconsistent depths (chair legs) correspond to more dispersed point cloud areas with low density (few points) in a neighborhood, indicating high uncertainty. In contrast, areas with accurate depths (chair seats) correspond to the points that are densely distributed on the true surface, indicating low uncertainty.}
    \vspace{-0.6cm}
    \label{fig:method-uncertainty}
\end{figure}

\noindent\textbf{Uncertainty Estimation.} 
Based on the observation that consistent depth will assemble back-projected points from different views tighter, leading to more certain points, we use the point density in a 3D neighborhood as the uncertainty. This is also a classic idea in point cloud denoising~\cite{zaman2017density,luo2021score}. To this end, we first back-project the monocular depths of each segmented instance from multi-view into world coordinate 3D space to form a point cloud, where the monocular depths are pre-aligned with the rendering depths through scale-shift invariant affine~\cite{yu2022monosdf}. We observe that the accurate depth points consistently fall on the surface of the instance. In contrast, the noisy points coming from inaccurate predictions are independently distributed along various viewing directions towards the object, thus exhibiting anisotropic distributions with large variance, as illustrated in Fig.~\ref{fig:method-uncertainty}. 

To further evaluate the density, we first downsample the fused point cloud to a fixed number (30,000 in our experiments) to decouple the relationship between the number of the points and the viewpoints. For the segmentation of the instance in each frame, we then back-project the masked monocular depth into 3D points and use ball query~\cite{qi2017pointnet++} to calculate the density of each point in small neighborhood, as shown in Fig.~\ref{fig:method-uncertainty}. The radius for ball query is defined as 
\vspace{-0.1cm}
\begin{equation}
    r=\text{Vol}(B_{opt}(P))+0.01,
\end{equation}
\vspace{-0.00cm}
where $P$ is the downsampled fused point cloud, $B_{opt}(P)$ denotes the minimum oriented bounding box of $P$~\cite{barequet2001efficiently} and $\text{Vol}$ denotes the volume of the bounding box. The densities are then normalized across all query points in all frames, 
\vspace{-0.1cm}
\begin{equation}
    d(p(u,v)) = \frac{d(p(u,v))}{\text{max}_{(u,v)\in \mathcal{S}_i}d(p(u,v))},
\end{equation}
\vspace{-0.00cm}
where $p(u,v)$ is the back-projected 3D point of pixel $(u,v)$, $d(p(u,v))$ is the measured density of that point and $\mathcal{S}_i$ is the segmented pixel area in the $i$-th image. The normalized densities are back-projected onto the image to obtain the per-pixel uncertainty estimation on the instance,
\vspace{-0.1cm}
\begin{equation}
    U_i(u,v) = 1-d(p(u,v)),
\end{equation}
\vspace{-0.00cm}
where $U_i(u,v)$ denotes the uncertainty at the pixel $(u,v)$ of the $i$-th image. We sequentially estimate the uncertainty for each instance in multi-view, thereby assembling complete uncertainty maps for all views.

\begin{figure}[t]
\vspace{-0.0cm}
  \includegraphics[width=\linewidth]{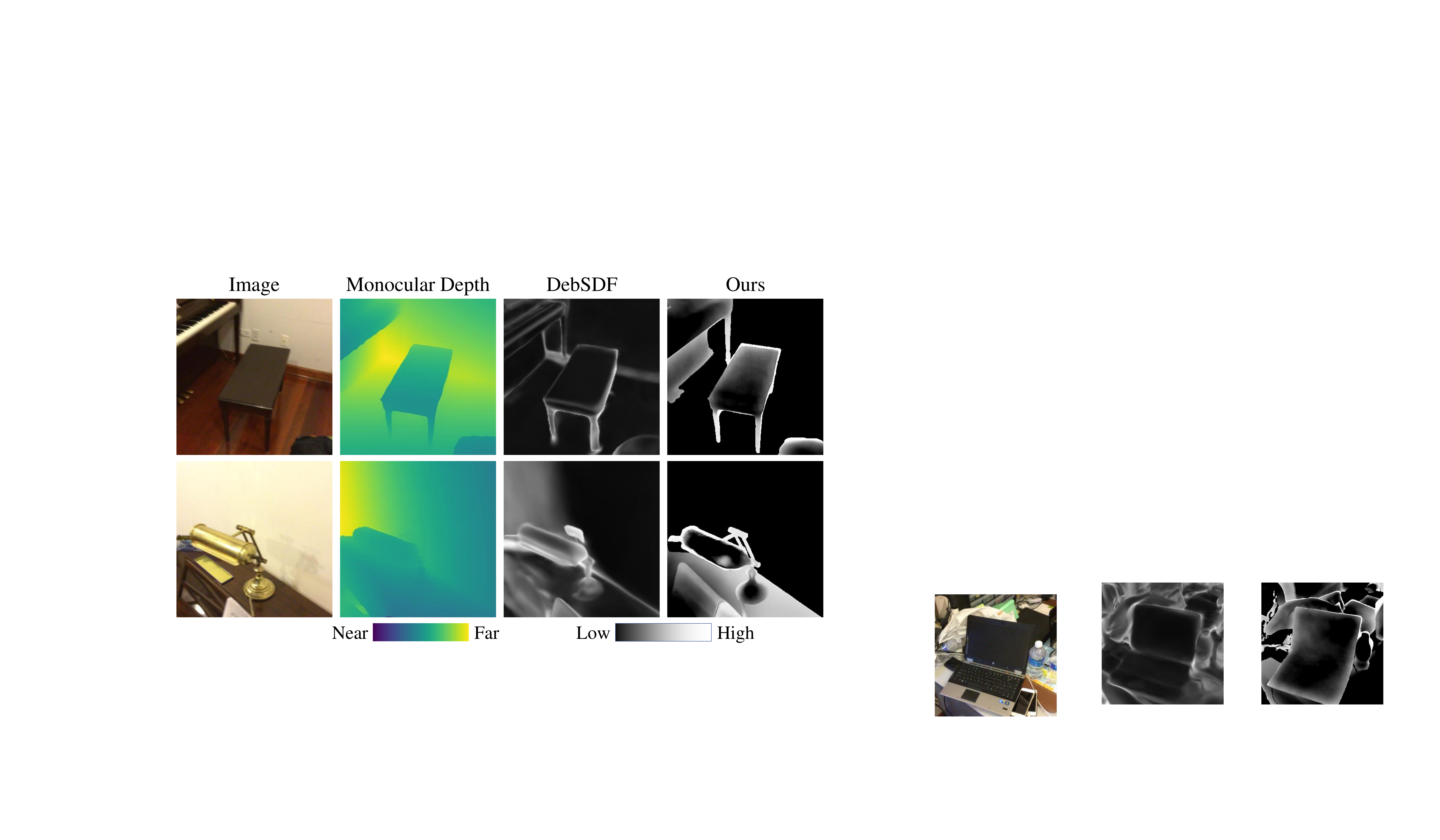}
  \vspace{-0.8cm}
    \caption{Visual comparison of the estimated uncertainty maps between DebSDF and ours. Our method estimates sharp uncertainty maps that faithfully capture the fine-grained geometric structures.}
    \vspace{-0.5cm}
    \label{fig:compare-uncertainty}
\end{figure}

\subsection{Adaptive Prior Loss and Uncertainty-Based Mask Constraint}
\label{sec:silhouette}

With the estimated uncertainty, we aim to reduce the negative impact of the inconsistency from the monocular clues and mine more reliable photo consistency as a remedy. First, we employ the estimated uncertainty maps as weights on the difference between monocular depths and the rendering ones, which filter out the impact brought by inaccurate supervision. This leads to an adaptive prior loss, as shown in Fig.~\ref{fig:overview}.

However, the regions of high-uncertainty, which often contain complex structures, are not effectively recovered by relying solely on photometric loss. To facilitate the learning of these areas, we further introduce an uncertainty-based instance mask constraint, enforcing the alignment of the learned instances within multi-view segmentation, as illustrated in Fig.~\ref{fig:overview}. Specifically, inspired by Pixel Warping~\cite{darmon2022neuralwarp}, for a ray emitted from a high-uncertainty instance region $S_r^i$ in the reference view $I_r$, we project points $\{p_j\}_{j=1}^K$ sampled on the ray into a nearby view $I_n$, and filter out the projected points $\{\pi_n(p_j)\}_{j=1}^K$ which fall within the instance mask $S_n^i$ in $I_n$. We then use the interpolated colors of these filtered projected points on $I_n$ and the corresponding predicted opacities $\alpha_j$ to render the final color,
\vspace{-0.2cm}
\begin{equation}
\label{eq:loss}
\begin{gathered}
    \hat{C}_n^{sil} = \sum_{j=1}^K \mathds{1}_j \cdot I_n[\pi_n(p_j)] \alpha_j \prod_{l<j}(1-\alpha_l), \\
    \mathds{1}_j = \begin{cases}
        1 & \pi_n(p_j)\in S_n^i \\
        0 & \pi_n(p_j)\notin S_n^i
    \end{cases}.
\end{gathered}
\end{equation}
The rendered color $\hat{C}_n^{sil}$ is compared with the corresponding ground truth color in $I_r$ as additional supervision. Unlike Pixel Warping~\cite{darmon2022neuralwarp}, we discriminately accumulate the projected points that just fall within the instance mask in the nearby view, because we are prompted of which sampling points contribute to the rendering of this instance through multi-view segmentation. This enables us to implicitly constrain these sampling points to align with the object surfaces.

\subsection{Optimization}
\label{sec:optimization}
\noindent\textbf{Uncertainty-Guided Ray Sampling.} We use the estimated uncertainty maps as probabilities to guide the ray sampling, paying more attention to regions with high uncertainty. We first allocate the number of sampling pixels for each instance according to its area in the segmentation. And then we calculate the sampling probabilities according to uncertainty.
The probability in $i$-th view is defined as $prob_i(u,v)=U_i(u,v)+0.05$, where the additional 0.05 ensures that the sampling is not omitted in areas with zero uncertainty.

\noindent\textbf{Training.} Our training process is divided into two stages. In the first stage, we uniformly apply monocular depth priors to learn a coarse representation of the scene. We then render low-resolution depth maps from all viewpoints to align the multi-view monocular depths to the same scale. Subsequently, we evaluate multi-view uncertainty for every segmented instance and assemble them to uncertainty maps of all frames. In the second stage, we integrate the uncertainty maps into the training process to utilize guided ray sampling, adaptive depth loss and instance mask constraints.

\noindent\textbf{Loss Function.} The overall loss function is defined as 
\vspace{-0.1cm}
\begin{equation}
    \mathcal{L}=\mathcal{L}_{color}+\lambda_1 \mathcal{L}_{eik}+\lambda_2 \mathcal{L}_{sil}+\lambda_3 \mathcal{L}_{d} +\lambda_4 \mathcal{L}_{n},
\end{equation}
where $\mathcal{L}_{eik}$ is the Eikonal term~\cite{yariv2020IDR}, $\mathcal{L}_{sil}$ is the instance mask constraint introduced in Sec.~\ref{sec:silhouette}, $\mathcal{L}_d$ is the adaptive depth loss and $\mathcal{L}_n$ is an optional adaptive normal loss. $\lambda_{1-4}$ are hyper-parameters for weighting each term.

\begin{table*}[t]
  \centering
  \caption{Averaged dense-view 3D reconstruction metrics on ScanNet and Replica datasets.}
  \vspace{-0.3cm}
  \label{tab:scannet-dense}
  \begin{adjustbox}{width=\linewidth,center}
  \begin{tabular}{l|ccccc|ccccc}
    \toprule
     \multirow{2}{*}{Methods} & \multicolumn{5}{c|}{ScanNet} & \multicolumn{5}{c}{Replica}  \\
    \cmidrule{2-11}
     & Acc$\downarrow$ & Comp$\downarrow$ & Prec$\uparrow$ & Recall$\uparrow$ & F-score$\uparrow$ & Acc$\downarrow$ & Comp$\downarrow$ & CD$\downarrow$ & N.C.$\uparrow$ & F-score$\uparrow$ \\
     \midrule
     UNISURF~\cite{oechsle2021unisurf}& 0.554 & 0.164 & 0.212 & 0.362 & 0.267 & 0.045 & 0.053 & 0.049 & 0.909 & 0.789 \\ 
     MonoSDF~\cite{yu2022monosdf} & 0.035 & 0.048 & 0.799 & 0.681 & 0.733 & 0.027 & 0.031 & 0.029 & 0.921 & 0.861 \\ 
     HybridNeRF~\cite{lyu2023learning} & 0.039 & 0.041 & 0.800 & 0.760 & 0.779 & 0.025 & 0.027 & \textbf{0.026} & 0.934 & \textbf{0.921} \\
     % Planar-SDF~\cite{zhou2024neural} & 0.053 & 0.056 & 0.714 & 0.664 & 0.688 & - & - & - & - & - \\ 
     H2O-SDF~\cite{park2023h2o} & \textbf{0.032} & 0.037 & 0.834 & 0.769 & 0.799 & - & - & - & - & - \\ 
     DebSDF~\cite{xiao2024debsdf} & 0.036 & 0.040 & 0.807 & 0.765 & 0.785 & 0.028 & 0.030 & 0.029 & 0.932 & 0.883 \\ 
     RS-Recon~\cite{yin2024ray} & 0.040 & 0.040 & 0.809 & 0.779 & 0.794 & 0.027 & \textbf{0.025} & \textbf{0.026} & 0.934 &  0.917 \\
     Ours & 0.035 & \textbf{0.032} & \textbf{0.846} & \textbf{0.824} & \textbf{0.834} & \textbf{0.024} & 0.029 & \textbf{0.026} & \textbf{0.937} & 0.918 \\ 
  \bottomrule
    \end{tabular}
    \end{adjustbox}
    \vspace{-0.2cm}
\end{table*}

\begin{figure*}[t]
\vspace{-0.0cm}
  \includegraphics[width=\linewidth]{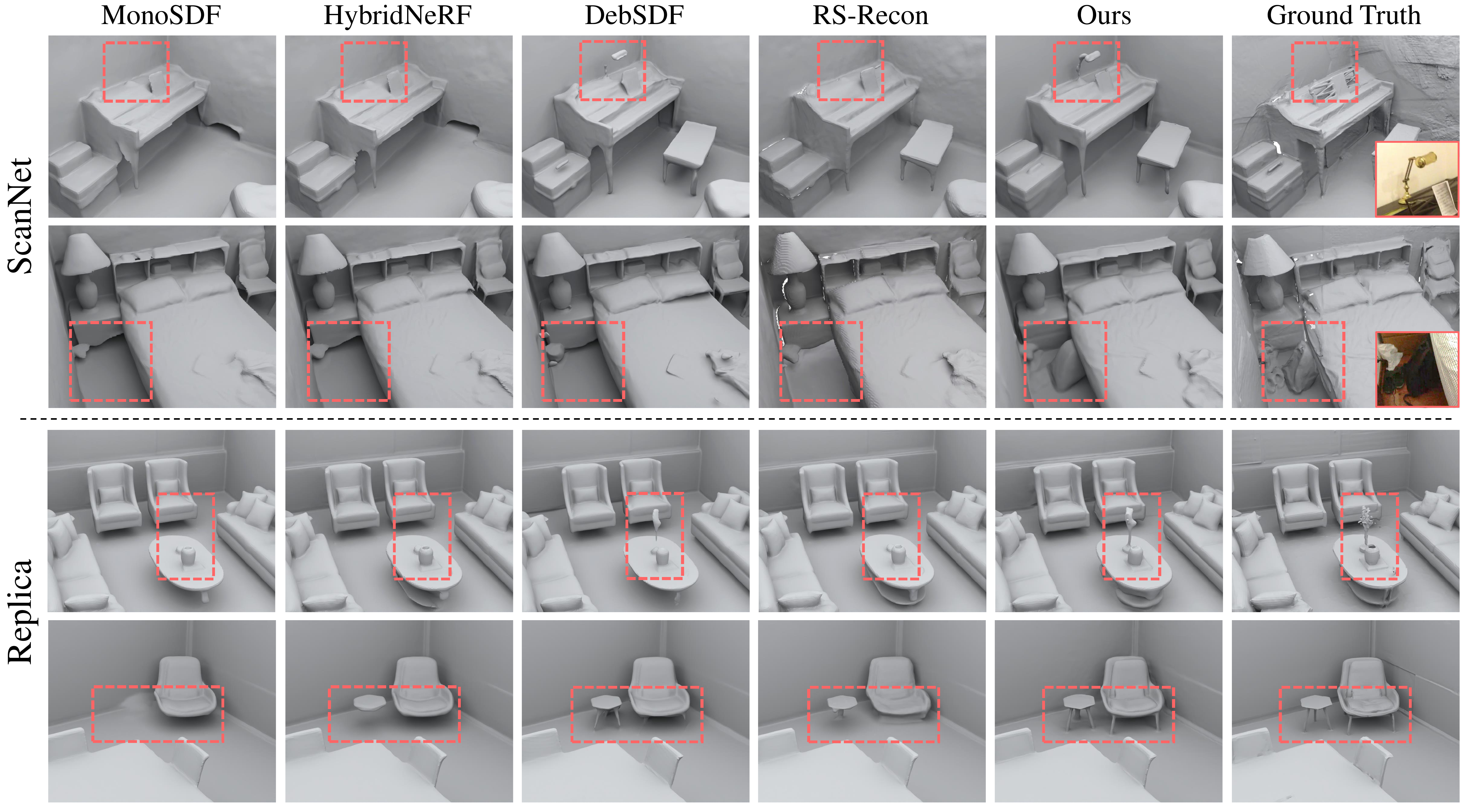}
  \vspace{-0.8cm}
    \caption{Visual comparisons of dense-view 3D reconstruction on ScanNet and Replica dataset.}
    \vspace{-0.5cm}
    \label{fig:denserecon-scannet-replica}
\end{figure*}

\section{Experiments}

To evaluate the effectiveness of our method, we conduct experiments based on various neural representation learning frameworks using multi-view images, including dense-view 3D reconstruction, sparse-view 3D reconstruction and sparse view synthesis.

\subsection{Dense-view 3D Reconstruction}

\noindent\textbf{Datasets.} We evaluate our performance under two real-world indoor scene datasets, including ScanNet~\cite{dai2017scannet} and Replica~\cite{straub2019replica}. We select 4 scenes from ScanNet and all 8 scenes from Replica, following baseline settings~\cite{yu2022monosdf, xiao2024debsdf}. Each scene consists of various numbers of observations from dense viewpoints, ranging from 200 to 400.

\noindent\textbf{Baselines and metrics.} We compare our method with the latest indoor scene reconstruction methods including MonoSDF~\cite{yu2022monosdf}, SDF-OCC-Hybrid~\cite{lyu2023learning} (shorted for ``HybridNeRF''), H2O-SDF~\cite{park2023h2o}, DebSDF~\cite{xiao2024debsdf}, RS-Recon~\cite{yin2024ray}. Note that the source code of H2O-SDF has not been made publicly available, thus we are unable to obtain its results on Replica dataset. Following baselines~\cite{yu2022monosdf,yin2024ray}, we report Chamfer Distance (CD), F-score in ScanNet dataset and additional Normal Consistency (N.C.) in Replica dataset.

\noindent\textbf{Implementation details.} We build our code upon the source code of MonoSDF~\cite{yu2022monosdf}. The hyper-parameters in Eq.~(\ref{eq:loss}) are set as $\lambda_1=0.1, \lambda_2=0.4, \lambda_3=0.5, \lambda_4=0.05$. Since the monocular normals are homologous with depths which come from the same foundation model, they show similar performances in the same regions of the images. Therefore, we can uniformly utilize the estimated uncertainty map to depth and normal priors. The nearby views used in Sec.~\ref{sec:silhouette} are selected according to the difference between observation angles. More implementation details are discussed in the supplementary materials.

\noindent\textbf{Comparisons.} We report numerical comparisons on ScanNet and Replica datasets in Tab.~\ref{tab:scannet-dense}. Our method outperforms all baseline methods on ScanNet dataset and achieves the highest normal consistency on Replica dataset. Visual comparisons in Fig.~\ref{fig:denserecon-scannet-replica} show that our method is capable of reconstructing fine-grained details of the scene, especially in the small thin structures such as the lamp on the piano, the flowers on the tea table and the chair legs.

\begin{table*}[t]
  \centering
  \caption{Averaged Chamfer Distance (CD) over the 15 scenes on DTU dataset in reconstructions from sparse views (small overlaps). NeuSurf$^\dag$ means NeuSurf with additional monocular cues.}
  \vspace{-0.3cm}
  \label{tab:dtu-sparse}
  \begin{adjustbox}{width=\linewidth,center}
  \begin{tabular}{c|cccccccc}
    \toprule
     Methods & COLMAP~\cite{schonberger2016structure} & SparseNeuS$_{ft}$~\cite{long2022sparseneus} & VolRecon~\cite{ren2023volrecon} & ReTR~\cite{liang2024retr} & NeuSurf~\cite{huang2024neusurf} & NeuSurf$^\dag$~\cite{huang2024neusurf} & UFORecon~\cite{na2024uforecon} & Ours  \\
    \midrule
     CD $\downarrow$ & 2.61 & 3.34 & 3.02 & 2.65 & 1.35 & 1.30 & 1.43 & \textbf{1.18} \\
  \bottomrule
    \end{tabular}
    \end{adjustbox}
\vspace{-0.4cm}
\end{table*}

\begin{figure*}[t]
\vspace{-0.0cm}
  \includegraphics[width=\linewidth]{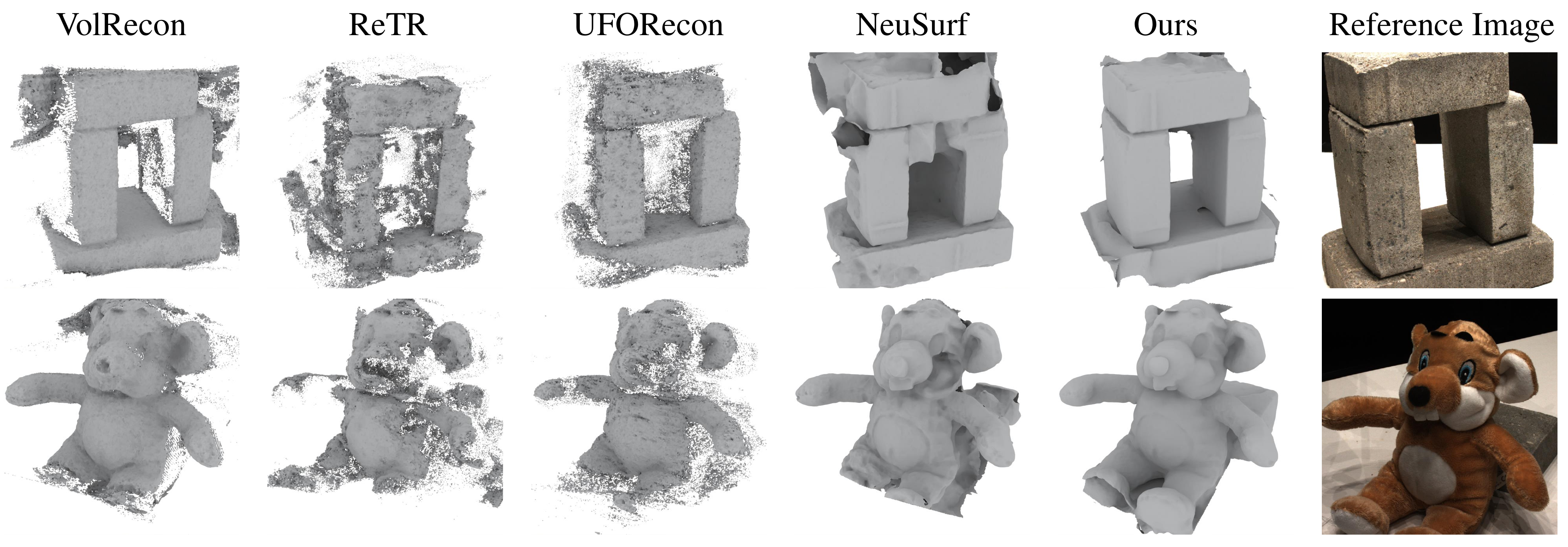}
  \vspace{-0.9cm}
    \caption{Visual comparisons on DTU dataset under the task of little-overlapping sparse input reconstruction.}
    \vspace{-0.5cm}
    \label{fig:sparserecon-dtu}
\end{figure*}

\subsection{Sparse-view 3D Reconstruction}

\noindent\textbf{Datasets.} We further evaluate our method in reconstructing 3D shapes from sparse observations on DTU dataset~\cite{jensen2014large}. Following previous methods~\cite{yu2020pixelnerf, huang2024neusurf}, we report our results on the 15 scenes, each of which shows single object with background from 3 viewpoints with small overlapping. 

\noindent\textbf{Baselines and metrics.} We compare our method with the latest sparse-view reconstruction approaches including the traditional MVS methods such as COLMAP~\cite{schonberger2016structure}, overfitting-based methods such as NeuSurf~\cite{huang2024neusurf}, generalizing-finetuning methods such as SparseNeuS~\cite{long2022sparseneus}, VolRecon~\cite{ren2023volrecon}, ReTR~\cite{liang2024retr} and UFORecon~\cite{na2024uforecon}. We use CD between the reconstructed meshes and the real-scanned point clouds as the evaluation metrics, following baselines~\cite{huang2024neusurf}.

\noindent\textbf{Implementation details.} We use the official code released by NeuSurf~\cite{huang2024neusurf} to produce our results of sparse-view reconstruction. The hyper-parameters in Eq.~(\ref{eq:loss}) are consistent with those employed in dense-view reconstruction. Since the multi-view images in each DTU scene capture the unique object, there is no need to conduct additional multi-view consistent instance segmentation. In our implementation, we first segment the scene into the object and the background, and then align and compute the uncertainty map only for the center object from various viewpoints.

\noindent\textbf{Comparisons.} We report numerical evaluations on DTU dataset in Tab.~\ref{tab:dtu-sparse}. For fair comparison, we also report the results of NeuSurf with monocular cues (NeuSurf$^\dag$), which are uniformly applied to all pixels, similar to MonoSDF~\cite{yu2022monosdf}. The superiority results in terms of CD show the effectiveness of our method. Further comparison between NeuSurf and NeuSurf$^\dag$ reveals that indiscriminately applying monocular depths to all pixels does not significantly improve the performance of NeuSurf. While our method leverages the estimated uncertainty maps to enhance the learning of the high-uncertainty regions, avoiding the misguidance from the inaccurate monocular priors. We showcase our improvements in visual comparison in Fig.~\ref{fig:sparserecon-dtu}, where our method consistently produces more complete and smoother surfaces compared to baseline methods. 

\begin{figure*}[t]
\vspace{-0.0cm}
  \includegraphics[width=\linewidth]{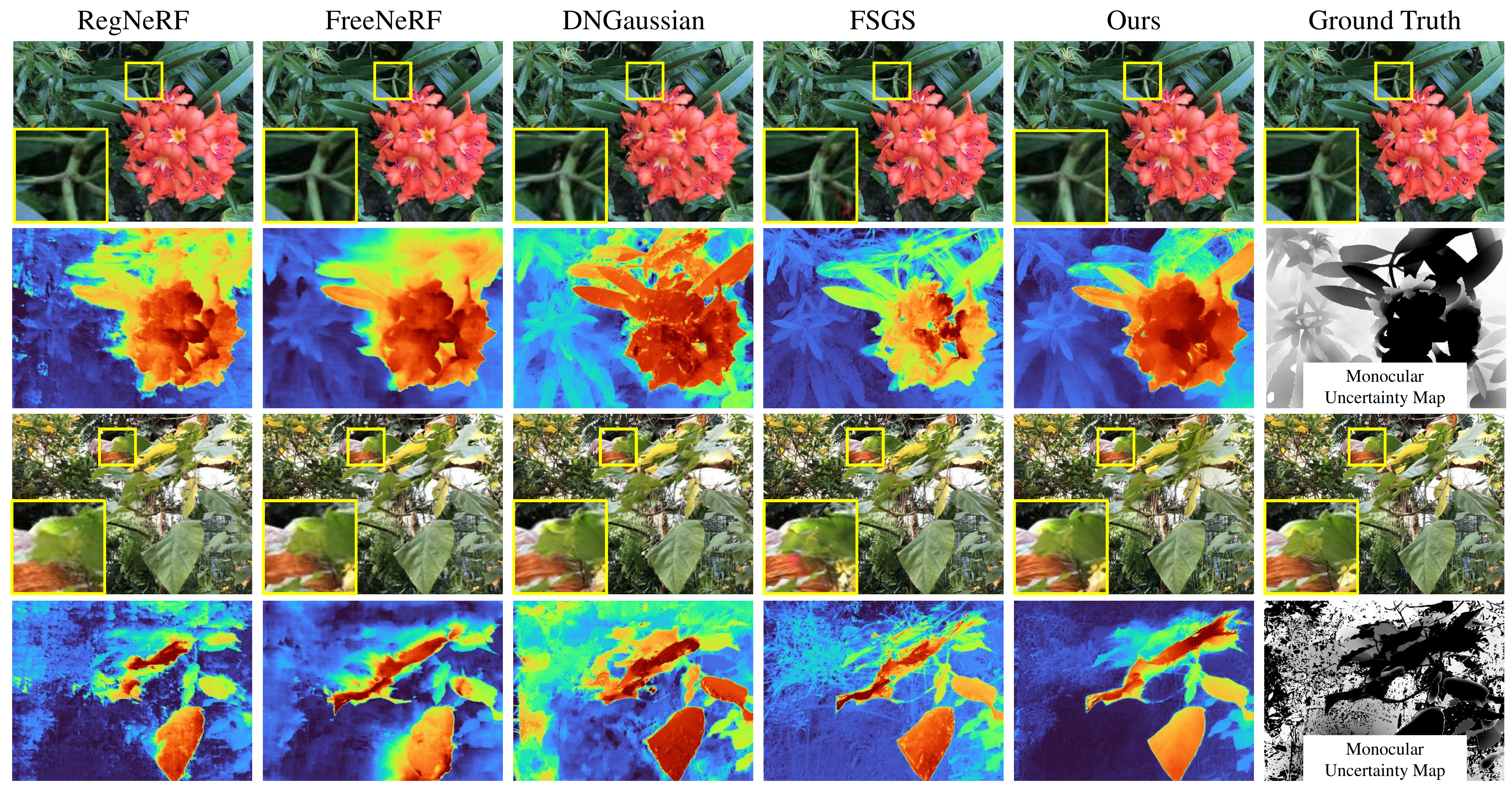}
  \vspace{-0.9cm}
    \caption{Visual comparisons of sparse novel view synthesis. In the uncertainty maps, areas that are more white indicate higher uncertainty.}
    \vspace{-0.7cm}
    \label{fig:sparserender-llff}
\end{figure*}

\begin{table}[t]
  \centering
  \caption{Quantitative comparison on LLFF dataset in novel view synthesis from sparse views.}
  \vspace{-0.3cm}
  \label{tab:llff-sparse}
  % \begin{adjustbox}{width=\linewidth,center}
  \begin{tabular}{l|ccc}
    \toprule
    Methods & PSNR$\uparrow$ & SSIM$\uparrow$ & LPIPS$\downarrow$  \\
    \midrule
    % DietNeRF~\cite{jain2021dietnetf} & 14.94 & 0.370 & 0.496 \\
    RegNeRF~\cite{niemeyer2022regnerf} & 19.08 & 0.587 & 0.336 \\
    FreeNeRF~\cite{yang2023freenerf} & 19.08 & 0.587 & 0.336 \\
    % SparseNeRF~\cite{wang2023sparsenerf} & 19.63 & 0.612 & 0.308 \\
    3DGS~\cite{kerbl20233dgs} & 15.52 & 0.405 & 0.408 \\
    DNGaussian~\cite{li2024dngaussian} & 19.12 & 0.591 & 0.294 \\
    FSGS~\cite{zhu2024fsgs} & 20.31 & 0.652 & 0.288 \\
    COR-GS~\cite{zhang2024corgs} & 20.45 & 0.712 & 0.196 \\
    Ours & \textbf{20.73} & \textbf{0.731} & \textbf{0.184} \\
  \bottomrule
    \end{tabular}
%     \end{adjustbox}
\vspace{-0.4cm}
\end{table}

\begin{figure}[t]
\vspace{-0.0cm}
  \includegraphics[width=\linewidth]{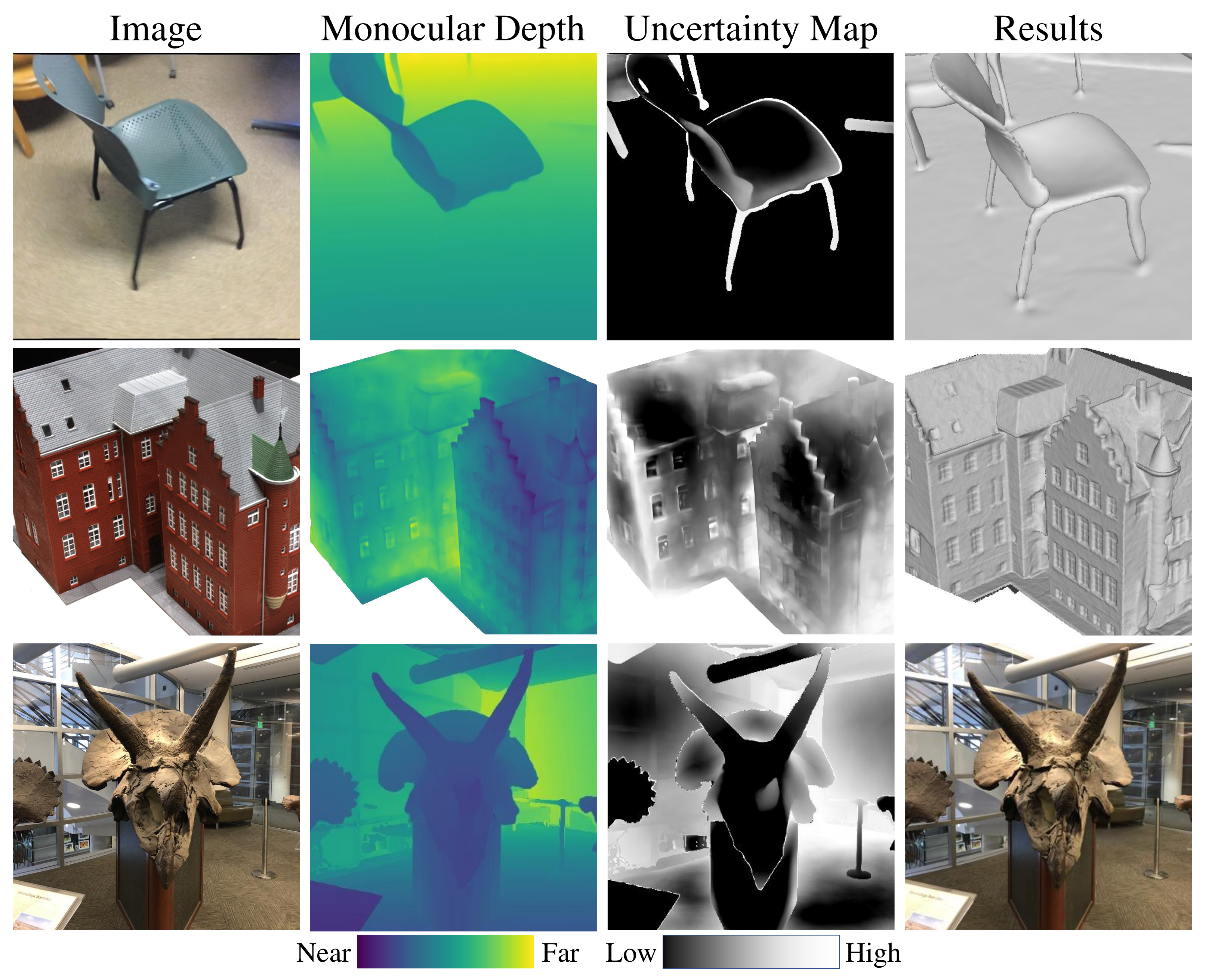}
  \vspace{-0.7cm}
    \caption{Visualization of our uncertainty maps calculated from monocular depths. Our uncertainties effectively identify the inconsistency across monocular clues on multi-view.}
    \vspace{-0.7cm}
    \label{fig:visual-uncertainty}
\end{figure}

\subsection{Sparse Novel View Synthesis}
\label{sec:subsec-nvs}

\noindent\textbf{Datasets.} We further evaluate our method on 3DGS-based sparse-input novel view synthesis (NVS) task on LLFF dataset~\cite{mildenhall2019local}. It contains 8 forward-facing real-world scenes. We select 3 views and downscale their resolutions as 8 to train, following previous works~\cite{zhu2024fsgs,niemeyer2022regnerf}. 

\noindent\textbf{Baselines and metrics.} We compare our method with latest few-shot NVS methods, including NeRF-based methods, such as RegNeRF~\cite{niemeyer2022regnerf}, FreeNeRF~\cite{yang2023freenerf} and 3DGS-based methods, such as DNGaussian~\cite{li2024dngaussian}, FSGS~\cite{zhu2024fsgs} and COR-GS~\cite{zhang2024corgs}. We report PSNR, SSIM~\cite{wang2004ssim} and LPIPS~\cite{zhang2018lpips} to evaluate the rendering quality following previous works~\cite{zhu2024fsgs,wang2023sparsenerf}. The implementation details of this section are provided in the supplementary materials.

% \noindent\textbf{Implementation details.} Our code in this experiment is built upon FSGS~\cite{zhu2024fsgs}, which utilizes monocular depths to enhance the rendering. $\mathcal{L}_{eik}$ and $\mathcal{L}_n$ in Eq.~(\ref{eq:loss}) are omitted in our experiment because there is no gradient fields in Gaussian splatting, and the orientation of 3D Gaussians are ambiguous during splatting~\cite{guedon2024sugar, gao2024relightable}. Note that 3D Gaussians are directly splatted onto the image plane with no sampled points in the space, thus we design a variant of our instance mask constraint, which encourages the projected instance depth points on the nearby view to move towards the mask of the same instance in nearby view, similar as~\cite{handrwr2020}.

\noindent\textbf{Comparisons.} The numerical and visual comparison are shown in Tab.~\ref{tab:llff-sparse} and Fig.~\ref{fig:sparserender-llff}. The visualizations of rendered images and depths further demonstrate our advanced results in recovering complex object details.
We further visualize our uncertainty maps across different datasets in Fig.~\ref{fig:visual-uncertainty}. Comparisons among the GT images, monocular depths, and the final results show that our method adaptively captures the inaccuracies in monocular depths, thereby achieving superior results beyond the quality of the priors.

\begin{table}[t]
  \centering
  \caption{Ablation study of each module on ScanNet dataset. Starting from the base model, we progressively add each of our module to reveal the impact of the proposed modules. }
  \vspace{-0.3cm}
  \label{tab:ablation-module}
  \begin{adjustbox}{width=\linewidth,center}
  \begin{tabular}{l|ccc}
    \toprule
     & Acc$\downarrow$ & Comp$\downarrow$ & F-score$\uparrow$  \\
    \midrule
    Base & 0.039 & 0.042 & 0.749 \\
    +Mono Uncertainty & 0.036 & 0.039 & 0.786 \\
    +Adaptive Sampling & 0.036 & 0.035 & 0.805 \\
    +Mask Constraint (Full) & \textbf{0.035} & \textbf{0.032} & \textbf{0.834} \\
  \bottomrule
    \end{tabular}
    \end{adjustbox}
\vspace{-0.2cm}
\end{table}

\begin{figure}[t]
\vspace{-0.0cm}
  \includegraphics[width=\linewidth]{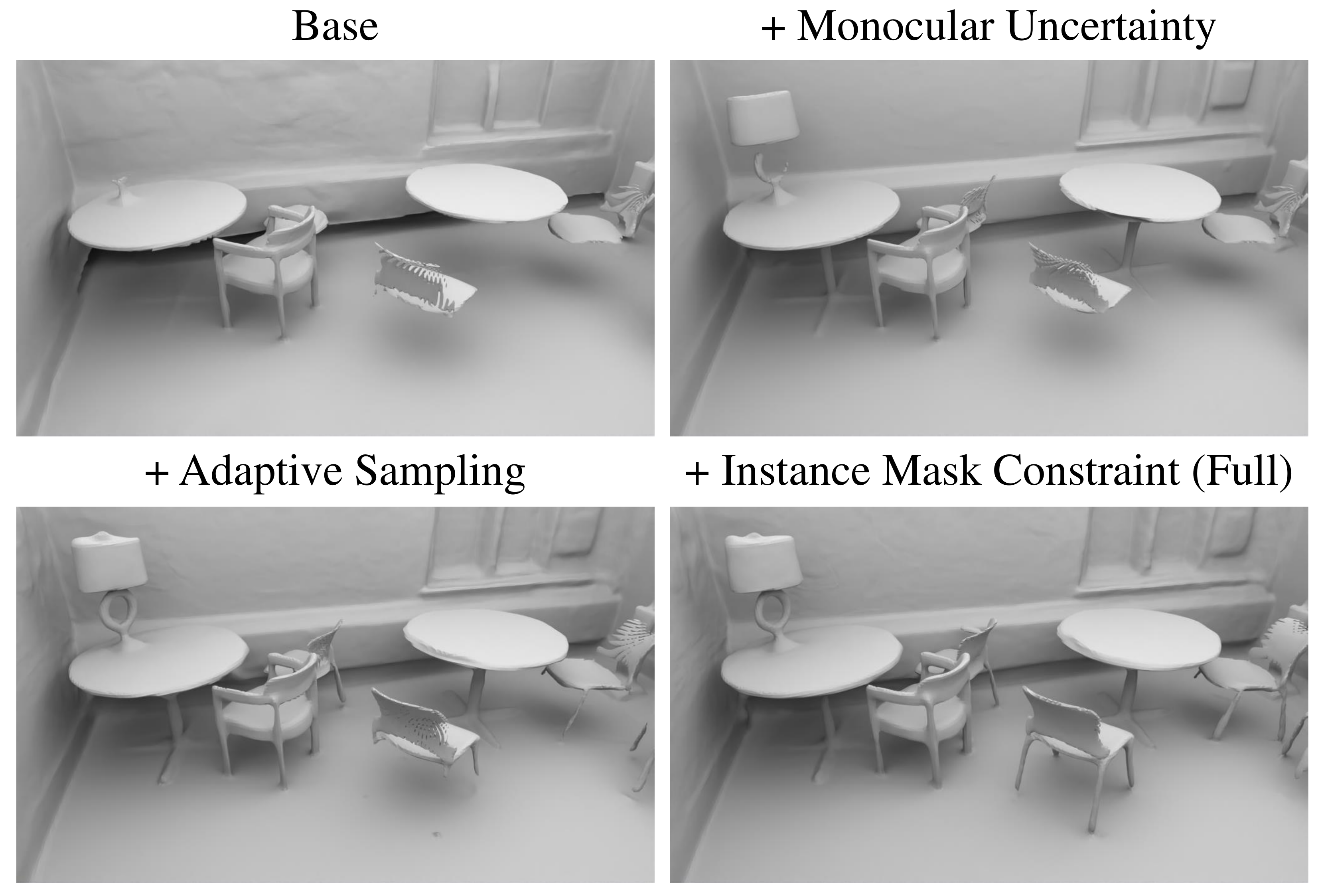}
  \vspace{-0.8cm}
    \caption{Visualization of ablations on each of our module.}
    \vspace{-0.5cm}
    \label{fig:ablation-module}
\end{figure}

\begin{table}[t]
  \centering
  \caption{Ablation study of different monocular priors. The results are averaged F-score across the four ScanNet scenes.}
  \vspace{-0.2cm}
  \label{tab:ablation-prior}
  \begin{adjustbox}{width=\linewidth,center}
  \begin{tabular}{l|ccc}
    \toprule
    Methods & Omnidata~\cite{eftekhar2021omnidata} & Metric3D v2~\cite{hu2024metric3dv2} & GeoWizard~\cite{fu2025geowizard}  \\
    \midrule
    MonoSDF & 0.733 & 0.749 & 0.741 \\
    Ours & 0.825 & 0.834 & 0.829 \\
  \bottomrule
    \end{tabular}
    \end{adjustbox}
\vspace{-0.5cm}
\end{table}

\subsection{Ablation Study}

\noindent\textbf{Effectiveness of each module.} We conduct ablation studies to justify the effectiveness of the modules in our method on ScanNet dataset.  Starting from the base model, which is identical to MonoSDF~\cite{yu2022monosdf}, we progressively add each of our modules to show the improvements of the reconstructed results. These additions include the adaptive monocular prior supervision, the uncertainty-guided ray sampling and the uncertainty-based instance mask constraint, as reported in Tab.~\ref{tab:ablation-module}. The visual comparisons in Fig.~\ref{fig:ablation-module} indicate that our method, equipped with each proposed module, successfully recovers complete and detailed geometric structures.

\noindent\textbf{Choice of monocular priors.} We further evaluate the performance of our method with different prior estimation models, including Omnidata~\cite{eftekhar2021omnidata}, Metric3D v2~\cite{hu2024metric3dv2} and GeoWizard~\cite{fu2025geowizard}. The improvement of our method beyond MonoSDF~\cite{yu2022monosdf} indicates that our method consistently enhances the monocular priors obtained from various estimation models. To fully reveal the potential of our approach, we choose Metric3D v2 as our primary prior model.

\section{Conclusion}

We propose MonoInstance, a novel approach to enhance monocular priors to provide robust monocular cues for multi-view neural rendering frameworks. To this end, we estimate the uncertainty of monocular priors by aligning multi-view instance depths in a unified 3D space and detecting the densities in point clouds. The estimated uncertainty maps are further utilized in adaptive prior loss, uncertainty-guided ray sampling and instance mask constraint. Our approach can be applied upon different multi-view neural rendering and reconstruction methods to enhance the monocular priors for better neural representation learning. Visual and numerical comparisons with the state-of-the-art methods justify our effectiveness and superiority over the latest methods.

\clearpage

\section{Acknowledgment}

We thank Ruihong Yin for generously providing their pre-trained meshes, as well as Shujuan Li and Junsheng Zhou for their valuable discussions and suggestions. This work was partially supported by Deep Earth Probe and Mineral Resources Exploration\text{—}National Science and Technology Major Project (2024ZD1003405), and the National Natural Science Foundation of China (62272263).

{
    \small
    \bibliographystyle{ieeenat_fullname}
    \bibliography{main}
}

\end{document}